%% file: main.tex
\documentclass{opt2020} 

\usepackage{cleveref}
\usepackage{bbm}
\usepackage{caption}
\usepackage{graphicx}
\usepackage{algorithm}

\title[Two-Level K-FAC Preconditioning for Deep Learning]{Two-Level K-FAC Preconditioning for Deep Learning}

\optauthor{\Name{Nikolaos Tselepidis} \Email{ntselepidis@student.ethz.ch}\\
\Name{Jonas Kohler} \Email{jonas.kohler@inf.ethz.ch}\\
\Name{Antonio Orvieto} \Email{antonio.orvieto@inf.ethz.ch}\\
\addr ETH Zurich, Switzerland}

\begin{document}

\maketitle

\begin{abstract}%
    \input{src/abstract.tex}
\end{abstract}


\section{Introduction} \label{sec:introduction}
\input{src/introduction.tex}

\section{Background on K-FAC Optimizer} \label{sec:background}
\input{src/background.tex}

\section{Two-Level K-FAC}  \label{sec:2lvl_kfac}
\input{src/2lvl_kfac.tex}

\section{Experimental Results} \label{sec:results}
\input{src/results.tex}

\section{Conclusion} \label{sec:conclusion}
\input{src/conclusion.tex}

\bibliography{main}
\clearpage
\appendix
\input{src/appendix}

\end{document}

%% file: src/abstract.tex
In the context of deep learning, many optimization methods use gradient covariance information in order to accelerate the convergence of Stochastic Gradient Descent.
In particular, starting with Adagrad~\cite{duchi2011adaptive}, a seemingly endless line of research advocates the use of diagonal approximations of the so-called empirical Fisher matrix in stochastic gradient-based algorithms, with the most prominent one arguably being Adam~\cite{kingma2014adam}.
However, in recent years, several works cast doubt on the theoretical basis of preconditioning with the empirical Fisher matrix~\cite{kunstner2019limitations, pascanu2013revisiting, martens2014new}, and it has been shown that more sophisticated approximations of the actual Fisher matrix more closely resemble the theoretically well-motivated Natural Gradient Descent~\cite{amari1998natural}.
One particularly successful variant of such methods is the so-called K-FAC optimizer~\cite{martens2015optimizing}, which uses a Kronecker-factored block-diagonal Fisher approximation as preconditioner.
In this work, drawing inspiration from two-level domain decomposition methods used as preconditioners in the field of scientific computing, we extend K-FAC by enriching it with off-diagonal (i.e. global) curvature information in a computationally efficient way.
We achieve this by adding a coarse-space correction term to the preconditioner, which captures the global Fisher information matrix at a coarser scale.
We present a small set of experimental results suggesting improved convergence behaviour of our proposed method.

%% file: src/introduction.tex
The question of how to efficiently incorporate curvature information into neural network training has been a long-standing issue in machine learning research.
In theory, the use of Hessian information allows to effectively escape saddle points, and improves both local and global convergence rates when used in a regularized Newton framework~\cite{cartis2011adaptive}.
However, the obvious drawback of Newton-type methods is that the computation per update step scales unfavorably with the problem dimension $d$, which can be extremely large in modern deep learning architectures.
Although most proposed second-order methods employ Krylov subspace iterations to compute their updates (e.g.~\cite{xu2019newton,kohler2017sub,wang2019stochastic}), thus making use of efficiently computable Hessian-vector products~\cite{pearlmutter1994fast}, the worst-case per-iteration complexity still scales as $\mathcal{O}(d^2)$, which is prohibitively large compared to first-order methods.
As a result, second-order optimizers are usually much slower in terms of run-time when compared to first-order optimizers (see e.g.~\cite{adolphs2019ellipsoidal,xu2020second}).
The same bottleneck can be found in a concurrent line of research which proposes the use of generalizations of the Gauss-Newton matrix (GGN)~\cite{schraudolph2002fast} instead of the Hessian (e.g.~\cite{martens2010deep,chapelle2011improved,pascanu2013revisiting}).
These two matrices match asymptotically in zero-residual non-linear least squares problems~\cite{nocedal2006numerical}.
Furthermore, the GGN resembles the well-known Fisher information matrix, used in Natural Gradient Descent (NGD), in many neural network settings~\cite{amari1998natural,martens2014new}.
GGN-vector products are also computed in $\mathcal{O}(d)$~\cite{schraudolph2002fast}, but since again up to $d$ might be needed per iteration, the total per-step complexity remains $\mathcal{O}(d^2)$.

As a result of the high per-iteration costs, researchers have advocated the use of approximations of the Hessian~(or GGN) matrix with algorithms ranging from quasi-Newton~\cite{bollapragada2018progressive} to sketched Newton~\cite{pilanci2017newton} and diagonal approximations~\cite{lecun2012efficient}\footnote{In some sense, most of the well known adaptive gradient methods such as RMSprop and Adam, can be ranked among such methods, but one must note that the applied diagonal preconditioners are approximations of the \textit{empirical} Fisher which may differ strongly from the real Fisher~\cite{kunstner2019limitations}.}.
The arguably most sophisticated and at the same time most performant method that emerged from this line of research is the so-called K-FAC algorithm~\cite{martens2015optimizing}.
This method makes use of a block-diagonal approximation of the Fisher matrix, which can be used very efficiently within Levenberg-Marquardt schemes, thanks to a Kronecker-factored form that allows inexpensive matrix inversion.
K-FAC, and some recent variants such as~\cite{george2018fast}, have been applied successfully for training all kinds of neural networks from Autoencoders~\cite{martens2015optimizing} over ResNets and CNNs~\cite{grosse2016kronecker} to RNNs~\cite{martens2018kronecker} and even transformers~\cite{zhang2019algorithmic}.
The block-diagonal K-FAC approximation has been shown to achieve a good balance between quality of curvature approximation and computational work.
As a result, the use of K-FAC preconditioner not only speeds up SGD in terms of iterations, but also in terms of wall-clock time, which is arguably what matters most to practitioners~\cite{zhang2019algorithmic}.
Nonetheless, the curvature signaled by K-FAC is missing any off-diagonal (cross-layer) information, which suggests possible improvements, especially for very deep networks.

To overcome this drawback, we propose a method for incorporating cross-layer information to the block-diagonal K-FAC approximation.
Our approach is inspired from coarse-grid correction techniques that have been widely used in the field of scientific computing~\cite{tang2009comparison,dolean2012analysis,napov2012algebraic,jolivet2014scalable}.
These methods have been shown to improve the convergence behaviour of domain decomposition preconditioners (i.e. block-diagonal approximations) when the number of subdomains (i.e. blocks) is increased~\cite{tang2009comparison,dolean2012analysis,napov2012algebraic,jolivet2014scalable}.
Based on this idea, we introduce a second level to the existing block-diagonal approximation that can effectively capture cross-layer information at a coarser scale. 

In Section \ref{sec:background}, we give a brief overview of Natural Gradient Descent (NGD) and K-FAC optimizer. 
In Section \ref{sec:2lvl_kfac}, we present our two-level approach for enriching K-FAC with off-diagonal covariance information, along with implementation details.
In Section \ref{sec:results}, we present preliminary experimental results, showing that capturing global covariance information at a coarse scale can indeed improve the convergence of K-FAC in very deep networks.

%% file: src/background.tex
\paragraph{Natural Gradient Descent.}
While standard Gradient Descent (GD) follows the directions of steepest descent in the parameter space, Natural Gradient Descent (NGD) preconditions the gradient using the Fisher information matrix in order to proceed along the directions of the steepest descent in the distribution space, with metric induced by the KL divergence.
More precisely, given tuples of labeled data $(x,y)\sim Q_{x,y}$, as well as an underlying parametric probabilistic model $p(y|x,\theta)$ (e.g. a neural network), the Fisher information matrix~(often denoted by $F$, for a formal definition see~\cite{martens2010deep}) describes the local curvature of the KL divergence between $p(x,y|\theta)$ and $p(x,y|\theta+\delta)$, in the sense that $\text{KL}(p(x,y|\theta)|p(x,y|\theta+\delta))=\frac{1}{2}\delta^\intercal F\delta + \mathcal{O}(\delta^3)$.
NGD can be written as an iterative procedure with the update rule: $\theta^{t+1} \leftarrow \theta^t - \eta F^{-1} \nabla_{\theta} \mathcal{L}(\theta^t)$, where $\theta^t$ are the model parameters at iteration $t$, $\mathcal{L}$ is the objective we aim to optimize, and $\eta>0$ is the learning rate.
\vspace{-2mm}
\paragraph{NGD and K-FAC as Second-Order Methods.}
Interestingly, the Fisher information matrix coincides with the Generalized Gauss-Newton (GGN) matrix in many neural network settings~\cite{martens2014new,pascanu2013revisiting}.
Since the GGN can be seen as a positive definite approximation of the Hessian\footnote{In fact, it is the Hessian of the local linearization of $\mathcal{L}$ as populated in the NTK literature~\cite{jacot2018neural}.
Furthermore, it has been shown that in networks with piecewise linear activations, the GGN and Hessian agree on the diagonal blocks~\cite{botev2017practical}.}, K-FAC is often regarded as a second-order algorithm that leverages curvature information.
In this regard, K-FAC computes an estimate of the natural gradient $F^{-1} \nabla \mathcal{L}$, using a block-diagonal approximation $\hat{F}$ of the Fisher $F$, where each block can be expressed as the Kronecker product of two factors of reduced order.
This factorization leads to substantial savings in computation and memory, and thus yields a highly efficient approximate second-order method with inherent parallelism~\cite{martens2015optimizing,ba2016distributed}.

\subsection{K-FAC on Neural Networks}
\label{sec:kfac_net}
Let us now consider the case of training a feed-forward neural network using K-FAC.
\vspace{-2mm}
\paragraph{Notation.}
Such a network is a function $f : \mathbb{R}^{d_0} \rightarrow \mathbb{R}^{d_L},$ parametrized by $\theta$, that maps a given input $a_0 \in \mathbb{R}^{d_0}$ to an output $a_L \in \mathbb{R}^{d_L}$, through a sequence of affine, and element-wise non-linear transformations.
In compact form, a feed-forward neural network can be written as $f(x,\theta) = W_L \cdot \varphi_{L-1} \left( W_{L-1} \cdot \varphi_{L-2} \left (\dots W_2 \cdot \varphi_1 \left( W_1 \cdot x \right) \right) \dots \right)$, where $W_i \in \mathbb{R}^{d_i \times d_{i-1}}, i = 1,\ldots,L$, denote the affine maps, and $\varphi_i$ are the element-wise non-linearities, also termed as activations.
For a given layer $i$, we denote its pre- and post-activations by $s_i = W_i \cdot \bar{a}_{i-1}$ and $a_i = \varphi_i \left( s_i \right )$, respectively.
$\bar{a}_i$ is formed by appending to $a_i$ an homogeneous coordinate with value one, so that every affine transformation can be expressed as a single matrix-vector product.
We gather all model parameters in a vector $\theta$, which is defined as $\theta=\left[\text{vec}(W_1)^\intercal,\text{vec}(W_2)^\intercal,\dots,\text{vec}(W_L)^\intercal \right]^{\intercal}$.
As usual, the operator $\text{vec}(\cdot)$ vectorizes matrices by stacking their columns together.
Moreover, let $\mathcal{L}$ denote a loss function of the form $\mathcal{L} \left(y, f(x, \theta)\right) = -\log{ r( y | f(x, \theta)) }$, which is associated with a predictive distribution $R_{ y | f(x, \theta) }:=P_{y|x}(\theta)$
used at the model's output, as well as the model distribution $P_{x,y}(\theta)$, with $r$ and $p$ being the respective probability density functions.
Examples of such functions are the standard least-squares-, as well as the cross-entropy loss~(for a proof, see~\cite{martens2014new}).
\vspace{-2mm}
\paragraph{Kronecker-Factored Approximate Fisher.}
Let us denote with $\mathcal{D} v$ the gradient of the loss $\mathcal{L}$ with respect to the quantity $v$.
The Fisher information matrix of a neural network parametrized by $\theta$ can be written as $F = E_{P_{y|x}(\theta),Q_x} \left[ \mathcal{D}\theta {\mathcal{D}\theta}^\intercal \right]$.
The layer-wise ordering of the parameters in $\theta$ induces an $L \times L$ block structure of the Fisher matrix $F$, where the $(i,j)$-th block is defined as
$F_{i,j} = E \left[ \text{vec}(\mathcal{D} W_i) \text{vec}(\mathcal{D} W_j)^\intercal \right] $.
Here, $D W_i$ is the gradient of $\mathcal{L}$ with respect to the weights $W_i$ of the $i$-th layer of the network, and
$g_i=\mathcal{D}s_i$ denotes the associated back-propagated loss derivatives, i.e. $\mathcal{D} W_i = g_i \bar{a}_{i-1}^\intercal$.
Using Kronecker products, it can be shown (see~\cite{martens2015optimizing}) that each Fisher block can be rewritten as:
\begin{equation}
F_{i,j} = 
E \left[ (\bar{a}_{i-1} \otimes g_i) (\bar{a}_{j-1} \otimes g_j)^\intercal \right] =
E \left[ (\bar{a}_{i-1} \otimes \bar{a}_{j-1}^\intercal) (g_i \otimes g_j^\intercal) \right].
\end{equation}
For the derivation of K-FAC, it is assumed that the products of the input activations are statistically independent with the products of the back-propagated derivatives~\cite{martens2015optimizing}.
Hence, $F_{i, j}$ is approximated by $\tilde{F}_{i,j}$ as follows:
\begin{equation}
\tilde{F}_{i,j} = 
E \left[\bar{a}_{i-1} \bar{a}_{j-1}^\intercal \right] \otimes E \left[ g_i g_j^\intercal \right]
= \bar{A}_{i-1, j-1} \otimes G_{i, j},
\label{eq:tilde_F}
\end{equation}
where $\bar{A}_{i,j} = E[\bar{a}_i \bar{a}_j^\intercal]$ and $G_{i,j} = E[g_i g_j^\intercal]$.
Next, in order to efficiently compute $\tilde{F}^{-1}$, K-FAC approximates $\tilde{F}$ either as block-diagonal or as block-tridiagonal, leading to two different variants~\cite{martens2015optimizing}.
In our work, we only consider the block-diagonal variant, which has attracted the interest of the deep learning community because of its inherent parallelism~\cite{ba2016distributed}.
In this case, the inverse of every diagonal block $\tilde{F}_{i,i}$ is computed as $\tilde{F}_{i,i}^{-1} = \bar{A}_{i-1, i-1}^{-1} \otimes G_{i,i}^{-1}$, without the need of explicitly forming and inverting $\tilde{F}_{i,i}$, hence reducing the computational work and memory requirements.

%% file: src/2lvl_kfac.tex
In order to compute an estimate of the natural gradient $F^{-1} \nabla \mathcal{L}$, the original one-level K-FAC utilizes a block-diagonal approximation $\hat{F}^{-1}$ of $\tilde{F}^{-1}$ (see equation~\eqref{eq:tilde_F}), and computes:
\begin{equation}
F^{-1} \nabla \mathcal{L} \approx
\hat{F}^{-1} \nabla \mathcal{L} = \text{diag}\left(\tilde{F}_{1,1}^{-1}, \tilde{F}_{2,2}^{-1}, \dots, \tilde{F}_{L,L}^{-1}\right) \nabla \mathcal{L}.
\label{eq:kfac_natural_grad}
\end{equation}
Approximating $\tilde{F}^{-1}$ as block-diagonal is equivalent to approximating $\tilde{F}$ as block-diagonal.
Therefore, the original one-level K-FAC utilizes only the intra-layer approximate covariances $\tilde{F}_{i,i}$, and ignores all off-diagonal blocks that represent the inter-layer covariances $\tilde{F}_{i, j}, \hspace{.1cm} i \neq j$.
In this section, motivated from this observation, we propose a method to incorporate inter-layer information into the one-level K-FAC preconditioner, in an attempt to improve convergence behaviour, especially for cases where cross-layer information is very important, such as (presumably) in very deep networks. 
\vspace{-3mm}
\paragraph{K-FAC as a Subspace Projection Method.}
Let us consider a neural network $f$ as in Section~\ref{sec:kfac_net}, with $L$ layers, and $n_i$ weights per layer, including the bias terms.
We denote the total number of parameters of the network with $n = \sum_{i=1}^{L} n_i$.
Let us consider the restriction matrices $V_i \in \{0, 1\}^{n_i \times n}$ that project a vector of $\mathbb{R}^{n}$ onto the layer $i, \ i = 1, \dots, L$, respectively.
Each restriction matrix $V_i$ is comprised of the subset of the rows $e_j$ of the identity matrix $I \in \mathbb{R}^{n \times n}$, where the $j$-th parameter belongs to the $i$-th layer of the network.
Using this notation, equation~\eqref{eq:kfac_natural_grad} can be rewritten as follows:
\begin{equation}
    F^{-1} \nabla \mathcal{L} \approx
    \hat{F}^{-1} \nabla \mathcal{L} = 
    \sum_{i = 1}^L V_i^\intercal \tilde{F}_{i,i}^{-1} V_i \nabla \mathcal{L}.
\end{equation}
It can be observed, that the one-level block-diagonal K-FAC preconditioner projects the gradient vector $\nabla \mathcal{L}$ on every layer independently, i.e. $y_i = V_i \nabla \mathcal{L}$, then scales the corresponding components of $y_i$ using local curvature information computed only from the intra-layer covariances, i.e. $\hat{y}_i = \tilde{F}_{i,i}^{-1} y_i$, and finally back-projects $\hat{y}_i$ onto the network $f$, combining the contributions from different layers to form the approximate natural gradient $\hat{F}^{-1} \nabla \mathcal{L}$.
\vspace{-3mm}
\paragraph{Enriching K-FAC with a Coarse-Space Correction.}
Let us now consider a restriction matrix $Z \in \{0, 1\}^{L \times n}$, that projects a vector of $\mathbb{R}^{n}$ onto the ``coarse'' network $f_{\text{coarse}}$ with $L$ layers and only a single weight per layer.
This matrix is defined as follows:
\begin{equation}
z_{i,j} = 
\begin{cases}
1 & \text{if the $j$-th component of $\theta$ belongs to the $i$-th layer of $f$}\\
0 & \text{otherwise}\\
\end{cases}.
\end{equation}
To the coarse network, we can associate the coarse representation of the Fisher, $F_{\text{coarse}} = Z F Z^\intercal \in \mathbb{R}^{L \times L}$, which captures the global covariance information at a coarse scale.
Since in K-FAC we consider the approximate Kronecker-factored Fisher $\tilde{F}$, we also introduce the associated approximation $\tilde{F}_{\text{coarse}} = Z \tilde{F} Z^\intercal \in \mathbb{R}^{L \times L}$, which equivalently can be rewritten as $[\tilde{F}_{\text{coarse}}]_{i,j} = \sum_{k,l} [\tilde{F}_{i,j}]_{k,l}$.

Based on this observation, we enrich K-FAC with an additional correction term, that operates on the global but coarse parameter space, capturing inter-layer covariance information, i.e.:
\vspace{-1mm}
\begin{equation}
    F^{-1} \nabla \mathcal{L} \approx
    \breve{F}^{-1} \nabla \mathcal{L} = 
    \sum_{i = 1}^L V_i^\intercal \tilde{F}_{i,i}^{-1} V_i \nabla \mathcal{L} + Z^\intercal \tilde{F}_{\text{coarse}}^{-1} Z \nabla \mathcal{L}.
\end{equation}
Thus, in order to compute an estimate of the natural gradient $F^{-1} \nabla \mathcal{L}$, our two-level approach additionally projects the gradient vector $\nabla \mathcal{L}$ onto the space associated with the coarse network $f_{\text{coarse}}$, scales it using the inverse of the coarse but global covariance, and projects it back to the space of the fine network $f$, to shift the independently preconditioned gradients by a different scalar for each layer.
Intuitively, the component $\sum_{i=1}^L V_i^\intercal \tilde{F}_{i,i}^{-1} V_i$ can be seen as a smoother that eliminates the high frequencies of the error $||F^{-1} \nabla \mathcal{L} - \breve{F}^{-1} \nabla \mathcal{L}||$, while the component $Z^\intercal \tilde{F}_{\text{coarse}}^{-1} Z$ operates on a coarser level, capturing the global trend of the natural gradient at a low resolution, and thus effectively eliminating the low frequencies of the error, without substantially increasing the computational work.
In particular, it only requires the computation and inversion of an $L \times L$ matrix, whenever the preconditioner is updated.
Similar ideas have been widely used for designing scalable parallel preconditioned iterative methods for solving large sparse linear systems in the field of scientific computing~\cite{tang2009comparison,dolean2012analysis,napov2012algebraic,jolivet2014scalable}.
We give an efficient algorithm for computing the coarse Fisher matrix $\tilde{F}_{\text{coarse}}$, along with implementation details in Appendix~\ref{sec:implementation}.



%
%
%
%
%
%
%
%
%
%
%



%% file: src/results.tex
\begin{figure}[t]
    \centering
    \includegraphics[width=0.46\textwidth]{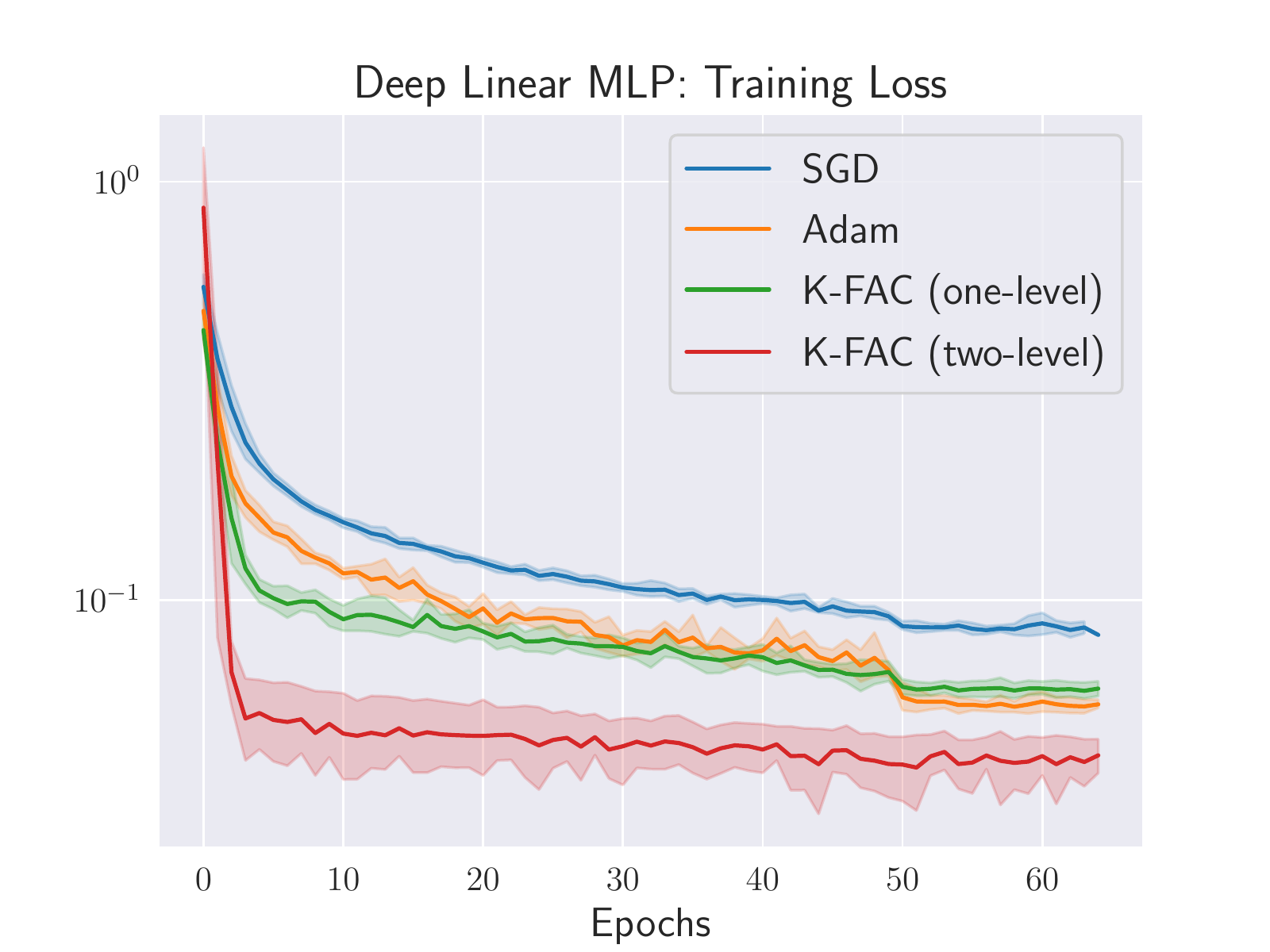}        
    \includegraphics[width=0.46\textwidth]{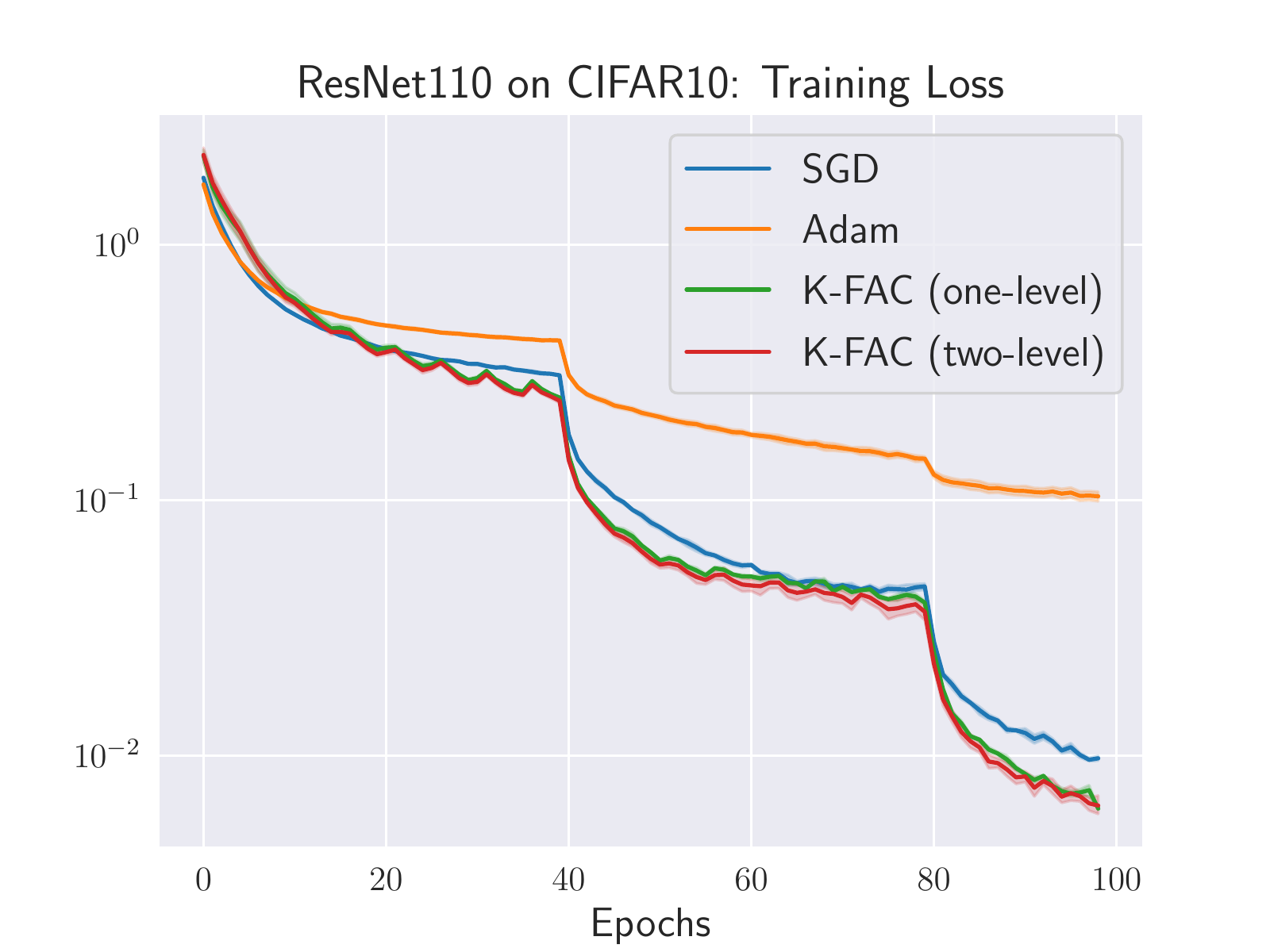}        
    \caption{Convergence of SGD, Adam, as well as one- and two-level K-FAC, when training a 64-layer linear MLP with planted targets (left), and ResNet110 on CIFAR10 (right). Mean and 95\% confidence interval of 5 independent runs.}
    \label{fig:train_loss}
\end{figure}

In this section, we present some preliminary empirical evidence that enriching K-FAC with inter-layer information, using a coarse-space correction term, can indeed improve the approximation quality of the preconditioner and lead to faster convergence. 
In particular, we compare our two-level approach with the standard one-level K-FAC in two neural network settings.
To put the results into perspective, we also benchmark SGD and Adam.
Regarding hyperparameters, we fixed the batch size for each optimizer and then grid-searched both learning rate ($\eta\in\{10^{-2},10^{-3},10^{-4}\}$) and momentum ($\mu\in\{0,0.9\}$).
Regarding the K-FAC variants, we tried the values $10^{-2}$ and $10^{-3}$ both for the damping parameter $\lambda$ and for the KL-clipping parameter $\kappa$.
We trained using the PyTorch framework~\cite{paszke2019pytorch} on a single NVIDIA GTX 1080 Ti with 11GB memory.

First, we trained a deep linear network with $64$ hidden layers ($10$ neurons per layer), and batch normalization, on five randomly generated datasets, with $10$-dimensional input samples drawn from a Gaussian distribution ($25k$ training and $2.5k$ test samples), and with binary targets coming from a separate randomly initialized one-layer linear network of the same width.
In this experiment, all optimizers operated on mini-batches of $512$ samples.
Further details on parameters can be found in Appendix~\ref{sec:experimental_setup}.
As can be seen in Fig.~\ref{fig:train_loss} (left), the proposed two-level K-FAC clearly outperforms its block-diagonal counterpart, as well as Adam and SGD.
In order to generalize this finding beyond simple linear networks, we also trained a ResNet110 on CIFAR10, using the cross-entropy loss.
Here, Adam and SGD were set to operate on batches of $64$ samples, while one- and two-level K-FAC used batch size $128$ (further details again in Appendix~\ref{sec:experimental_setup}).
As can be seen in Fig.~\ref{fig:train_loss} (right), the coarse-space correction still works quite well in this setting, but the margin to one-level K-FAC is significantly reduced compared to case of the deep linear MLP\footnote{At this point, we can only hypothesize, but one possible explanation could be that the network is highly over-parametrized for the simple task of CIFAR10 classification, which usually makes optimization much easier~\cite{arora2018optimization}.}.

In summary, our findings suggest that two-level K-FAC can indeed enhance convergence in deep neural networks, but further investigation is needed to identify settings where off-diagonal Hessian information is particularly useful.
We consider this to be an interesting direction of future research.

%% file: src/conclusion.tex
We proposed a two-level extension to K-FAC that incorporates a coarse-space correction term in order to efficiently capture the global structure of the Fisher information matrix and improve the convergence behaviour of the optimizer.
Our experiments show that the use of off-diagonal covariance information can indeed yield enhanced optimization performance of K-FAC in the case of (very) deep networks.
Going forward, we believe that the identification of more such settings, where cross-layer information is important to consider for optimizers, is an interesting direction of future research.
In particular, it is yet to be understood how advanced network architectures such as normalization layers, residual connections, and attention layers alter layer dependencies, and hence off-diagonal Hessian information.

%% file: src/appendix.tex
\section{Implementation Details} \label{sec:implementation}

\paragraph{Coarse Fisher Matrix Computation.}


The $(i,j)$-th element of $\tilde{F}_{\text{coarse}}$ is equal to the sum of all elements in the $(i,j)$-th block of $\tilde{F}$.
Moreover, every block $\tilde{F}_{i,j}$ is equal to $\tilde{F}_{i,j} = \bar{A}_{i-1,j-1} \otimes G_{i,j}$.
Explicitly computing $\tilde{F}_{i,j}$, and then summing up all the elements requires substantial computational work and excessive memory requirements.
To overcome this issue, we directly compute the required sum without explicitly forming $\tilde{F}_{i,j}$, as follows:
\begin{equation}
    \sum_{k,l} \left[ \tilde{F}_{i,j} \right]_{k,l} = 
    \sum_k \left[ \tilde{F}_{i,j} \cdot \mathbbm{1} \right]_{k} = 
    \sum_k \left[ \left(\bar{A}_{i-1,j-1} \otimes G_{i,j}\right) \cdot \mathbbm{1} \right]_{k} = 
    \sum_{k,l} \left[ G_{i,j} \cdot \mathbbm{1}_{m \times n} \cdot \bar{A}_{i-1,j-1} \right]_{k,l},
\label{eq:Fij_coarse}
\end{equation}
where $\mathbbm{1}$ is a vector of ones, and $\mathbbm{1}_{m \times n}$ is the same vector reshaped into an $m \times n$ matrix, where $m$ and $n$ are such that the dimensions match for the matrix multiplications.
\Cref{eq:Fij_coarse} provides an efficient way for computing every element of $\tilde{F}_{\text{coarse}}$, and since $\tilde{F}_{\text{coarse}}$ is symmetric (similarly to $\tilde{F}$), one only needs to compute its upper or lower triangular part.
The algorithm for the computation of the coarse Fisher matrix $\tilde{F}_{\text{coarse}}$ is given below:

\begin{algorithm2e}
\LinesNumbered
\SetKwInOut{Input}{in}
\SetKwInOut{Output}{out}
\Input{Input activations $\bar{a}_{i-1}^{(t)}$ and back-propagated gradients $g_i^{(t)}$ for $i = 1, \dots, L$, at iteration $t$.}
\Output{Coarse approximate Fisher information matrix $\tilde{F}_{\text{coarse}}$}
\Begin{
Set $\epsilon = \min{(1 - 1/t, 0.95)}$ \hfill \# statistical decay\;

\For{$i = 1, \dots, L$}{
\For{$j = 1, \dots, i$}{
Compute $\bar{A}_{i-1,j-1}^{(t)} = E\left[\bar{a}_{i-1}^{(t)} (\bar{a}_{j-1}^{(t)})^\intercal\right]$ \hfill\# downsample $\bar{a}_{i-1}^{(t)}$ or $\bar{a}_{j-1}^{(t)}$ if needed\;
Update $\bar{A}_{i-1,j-1} = \epsilon \bar{A}_{i-1,j-1} + (1 - \epsilon) \bar{A}_{i-1,j-1}^{(t)}$\;

Compute $G_{i,j}^{(t)} = E\left[g_i^{(t)} (g_j^{(t)})^\intercal\right]$ \hfill\hfill\# downsample $g_i^{(t)}$ or $g_j^{(t)}$ if needed\;
Update $G_{i,j} = \epsilon G_{i,j} + (1 - \epsilon) G_{i,j}^{(t)}$\;

    Compute $\left[ \tilde{F}_{\text{coarse}} \right]_{i,j} = \sum_{k,l} \left[ \tilde{F}_{i,j} \right]_{k,l}$ without forming $\bar{A}_{i-1, j-1} \otimes G_{i,j}$ and summing up all elements, but using the formula $\sum_{k,l} \left[ G_{i, j} \cdot \mathbbm{1}_{m\times n} \cdot \bar{A}_{i-1, j-1} \right]_{k,l}$, where $m$ and $n$ are such that the dimensions match;
}
}
$\tilde{F}_{\text{coarse}} = \tilde{F}_{\text{coarse}} + \tilde{F}_{\text{coarse}}^{\intercal} - \text{diag}(\tilde{F}_\text{coarse})$ \hfill\hfill \# here, diag$(\cdot)$ yields a diagonal matrix\;
}
\caption{Coarse Fisher Matrix Computation}
\end{algorithm2e}

It should be noted, that in the case of convolutional layers the dimensions of the input activations $\bar{a}_{i-1}^{(t)}$ and back-propagated gradients $g_i^{(t)}$, between different layers, i.e. for $i \neq j$, may not match, and thus the inter-layer covariances $\bar{A}_{i-1,j-1}^{(t)}$ and $G_{i,j}^{(t)}$ cannot be computed.
To tackle this issue, we downsample the feature maps of larger dimensions, so that we can compute the required covariances.
In our implementation, we use the nearest-neighbor downsampling algorithm as implemented in PyTorch.
Finally, we need to mention that we keep running estimates of the statistics $\bar{A}_{i-1,j-1}$ and $G_{i,j}$ as shown in lines 6 and 8 of Algorithm 1.

\paragraph{Damping.}

Although, in theory the block-diagonal K-FAC preconditioner can be inverted block-wise with the inverse of each block being $\tilde{F}_{i, i}^{-1} = \bar{A}_{i-1, i-1}^{-1} \otimes G_{i, i}^{-1}$, in practice, a damping term $\lambda I$ is usually added to every $\tilde{F}_{i,i}$ in order to account for the inaccuracies of the approximation and ill-conditioning of the diagonal blocks.
The addition of this term makes the use of the previous formula impossible.
\citet{martens2015optimizing} proposed two methods to resolve this issue; (i) an exact method that is based on the eigenvalue decomposition of the diagonal blocks, and (ii) an approximate but more computationally efficient approach, which they refer to as factored Tikhonov regularization.
In the latter case, every block $\tilde{F}_{i,i} = \bar{A}_{i-1, i-1} \otimes G_{i, i} + \lambda I$ is approximated by:
\begin{equation}
\tilde{F}_{i,i} \approx    
\left( \bar{A}_{i-1, i-1} + \pi_i \lambda^{1/2} I \right) 
\otimes \left( G_{i, i} + \frac{1}{\pi_i} \lambda^{1/2} I \right) \text{ where }
\pi_i = \sqrt{ \frac{ \text{tr}( \bar{A}_{i-1, i-1} ) / ( d_{i-1} + 1 ) }{ \text{tr}( G_{i,i} ) / d_i } }.
\end{equation}
Here, $d_i$ is the dimension (number of units) in layer $i$.
Therefore, using this approach as well as the formula $(A \otimes B) \text{vec}(X) = \text{vec}(B X A^\intercal)$, the $i$-th block of the natural gradient can be computed as:
\begin{equation}
(\tilde{F}_{i,i} + \lambda I)^{-1} \nabla \mathcal{L}_i = 
\text{vec}\left( \left( G_{i, i} + \frac{1}{\pi_i} \lambda^{1/2} I \right)^{-1} \nabla \mathcal{L}_i^{\text{reshaped}} \left( \bar{A}_{i-1, i-1} + \pi_i \lambda^{1/2} I \right)^{-1}\right ).
\end{equation}

Concerning the inversion of the coarse Fisher matrix $\tilde{F}_{\text{coarse}}$, we actually compute $(\tilde{F}_{\text{coarse}} + \lambda I)^{-1}$, 
where $\lambda$ is equal to the damping parameter used when inverting the diagonal blocks.

\paragraph{KL-clipping.}

After preconditioning the gradients, we scale them by a factor $\nu$ which is given by the equation:
\begin{equation}
\nu = \text{min}\left(1, \sqrt{\frac{\kappa}{\eta^2 \sum_{i=1}^n |\mathcal{G}_i^\intercal \nabla \mathcal{L}_i(\theta_i)|}}\right),
\end{equation}
where $\mathcal{G}$ is the preconditioned gradient, $\eta$ is the learning rate, and $\kappa$ is a user defined parameter.
We choose $\kappa$ so that the square Fisher norm is at most $\kappa$~\cite{martens2015optimizing}.

\section{Experiments} 
\subsection{Details on Settings and Parameters}\label{sec:experimental_setup}

\paragraph{Deep Linear MLP.}

In this experiment, all optimizers operated on mini-batches of $512$ samples, with learning rate $\eta = 10^{-3}$, momentum $\mu = 0.9$, and weight decay $\beta=10^{-3}$.
The one- and two-level K-FAC, were configured so that they update the running estimates of the covariances every $10$ iterations, and recompute the preconditioner every $100$ iterations.
Moreover, the damping parameter $\lambda$ was set to $10^{-2}$, and the parameter $\kappa$ used for KL-clipping was set to $10^{-3}$.
The inverses of the diagonal blocks in one- and two-level K-FAC were computed using eigen-decomposition~\cite{martens2015optimizing}, since it seemed to be more robust for the case of the deep linear MLP.

\paragraph{ResNet110.}

Similarly to the previous experiment, we configured all optimizers to use momentum with $\mu=0.9$ and weight decay with $\beta=10^{-3}$.
The learning rate was set to $10^{-2}$ for all optimizers except for Adam who seemed to work better with $10^{-3}$, since setting a larger learning rate led to a substantial increase in the test loss, reducing generalization.
Moreover, a learning rate schedule was chosen so that the learning rate $\eta$ is reduced by a factor of $10$, on epochs $40$ and $80$.
For the one- and two-level K-FAC the damping parameter $\lambda$ was set to $10^{-3}$, while the KL-clipping parameter $\kappa$ was set to $10^{-2}$.
It should be mentioned, that in order to keep the computational work at low levels, we configured one- and two-level K-FAC to update the running estimates of the covariances every $200$ iterations, and recompute the preconditioner every $2000$ iterations.
Moreover, the inverses of the diagonal blocks were computed using the factored Tikhonov regularization technique (see~\Cref{sec:implementation}).

\subsection{Additional Experimental Results} \label{sec:additional_results}

\begin{figure}[h]
    \centering
    \includegraphics[width=0.47\textwidth]{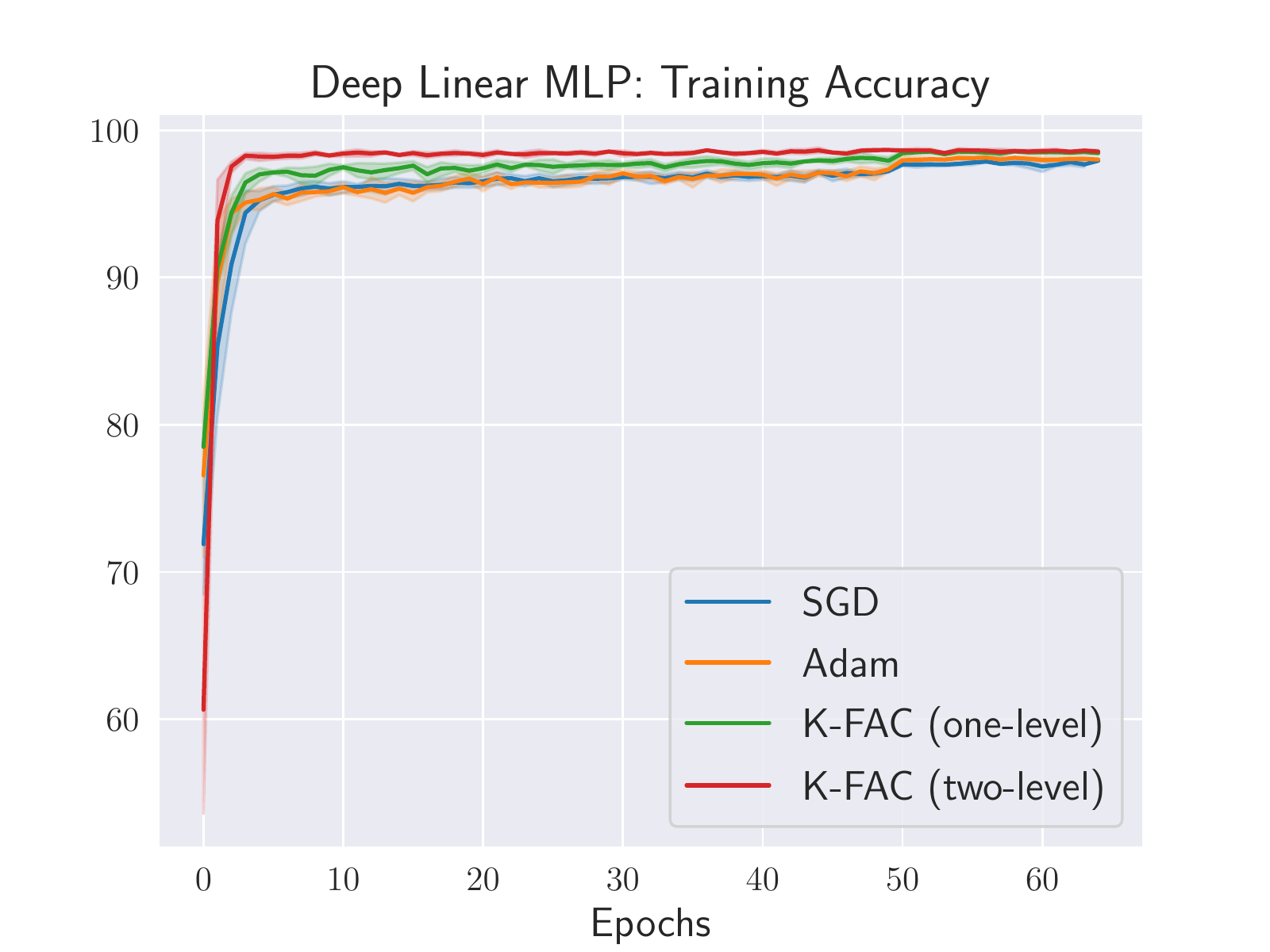}        
    \includegraphics[width=0.47\textwidth]{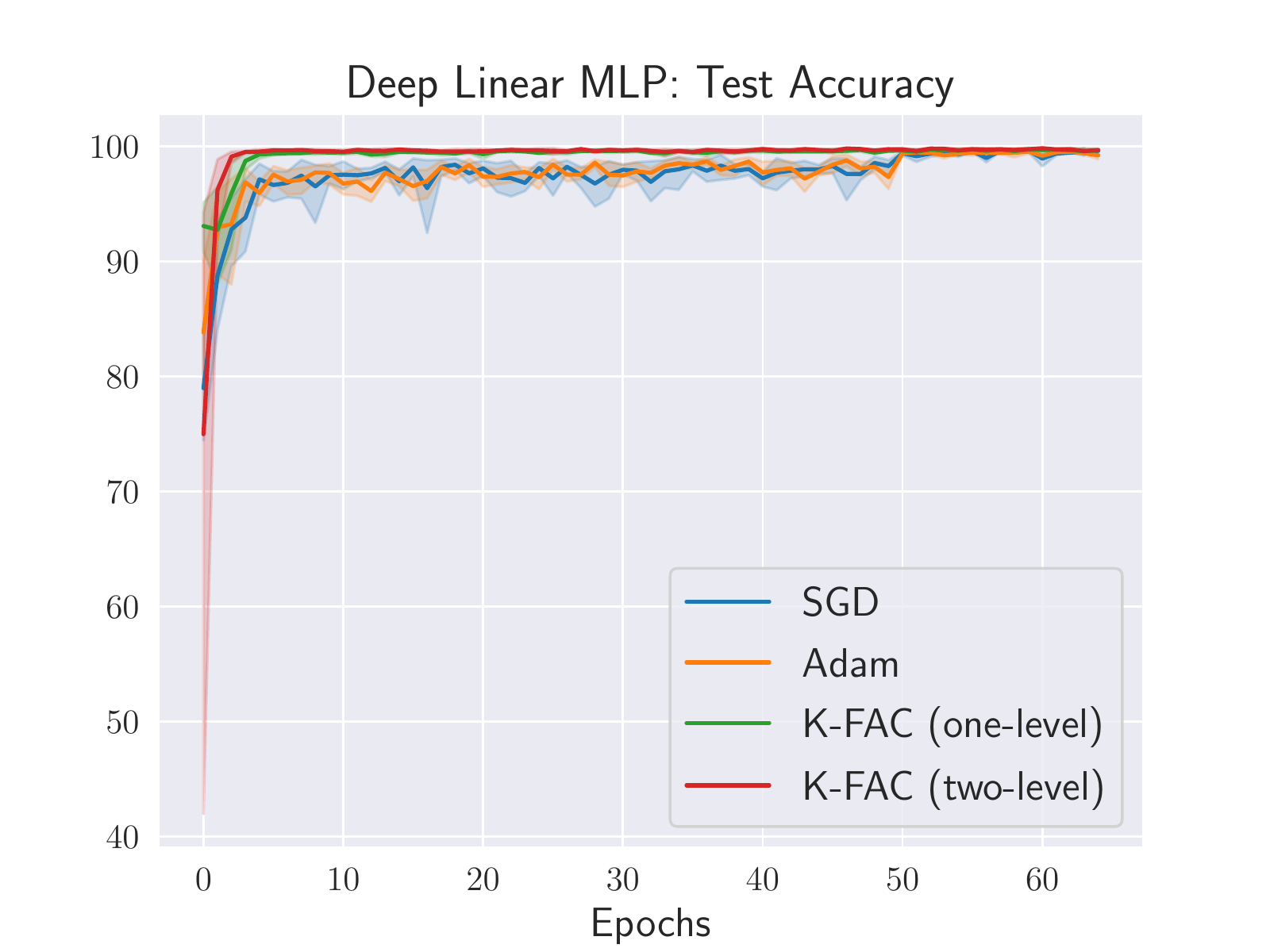}        
    \caption{Training and test accuracy per epoch for SGD, Adam, as well as one- and two-level K-FAC, when training a linear 64-layer MLP with planted targets.}
\end{figure}

\begin{figure}[h]
    \centering
    \includegraphics[width=0.47\textwidth]{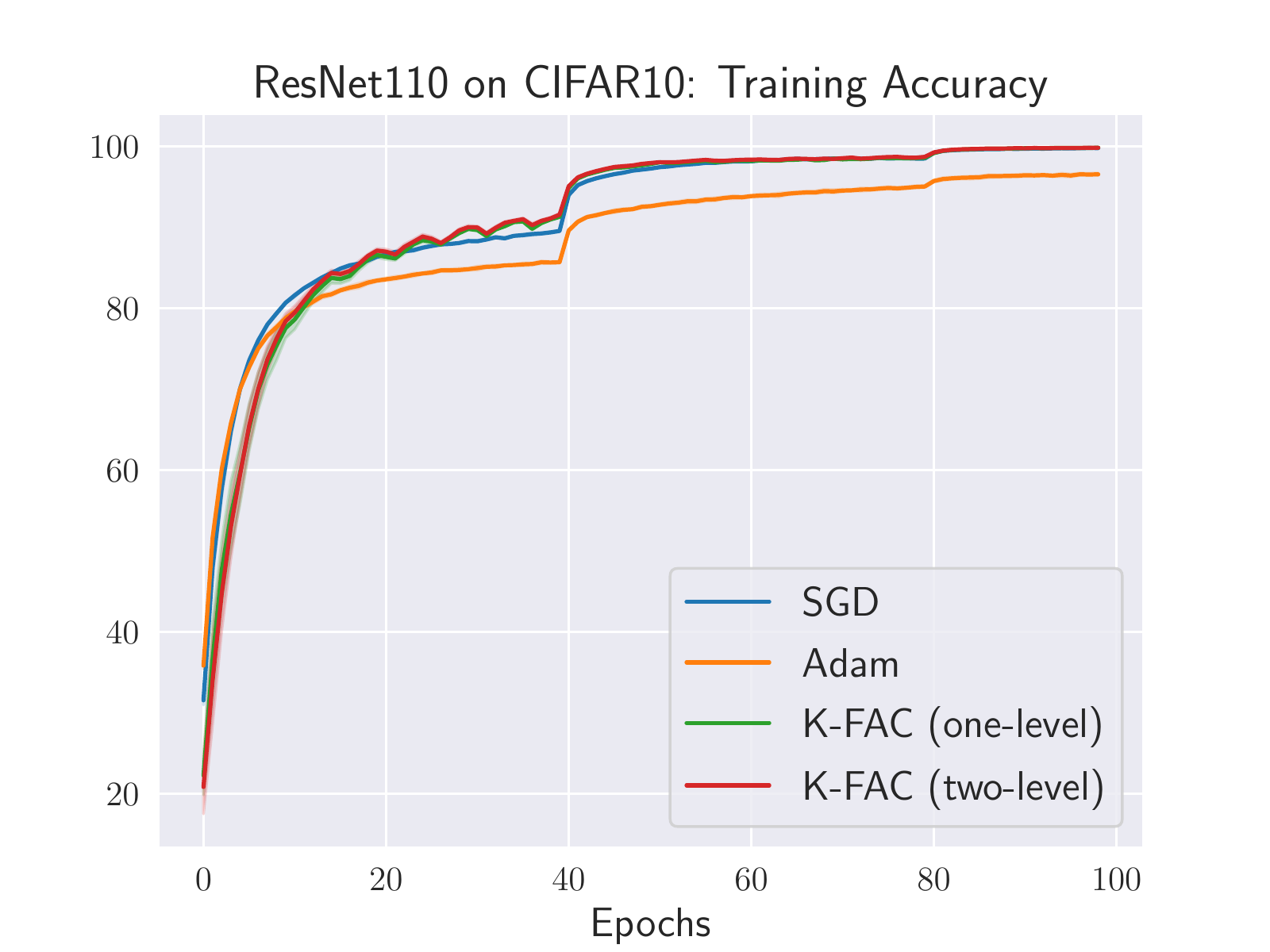}
    \includegraphics[width=0.47\textwidth]{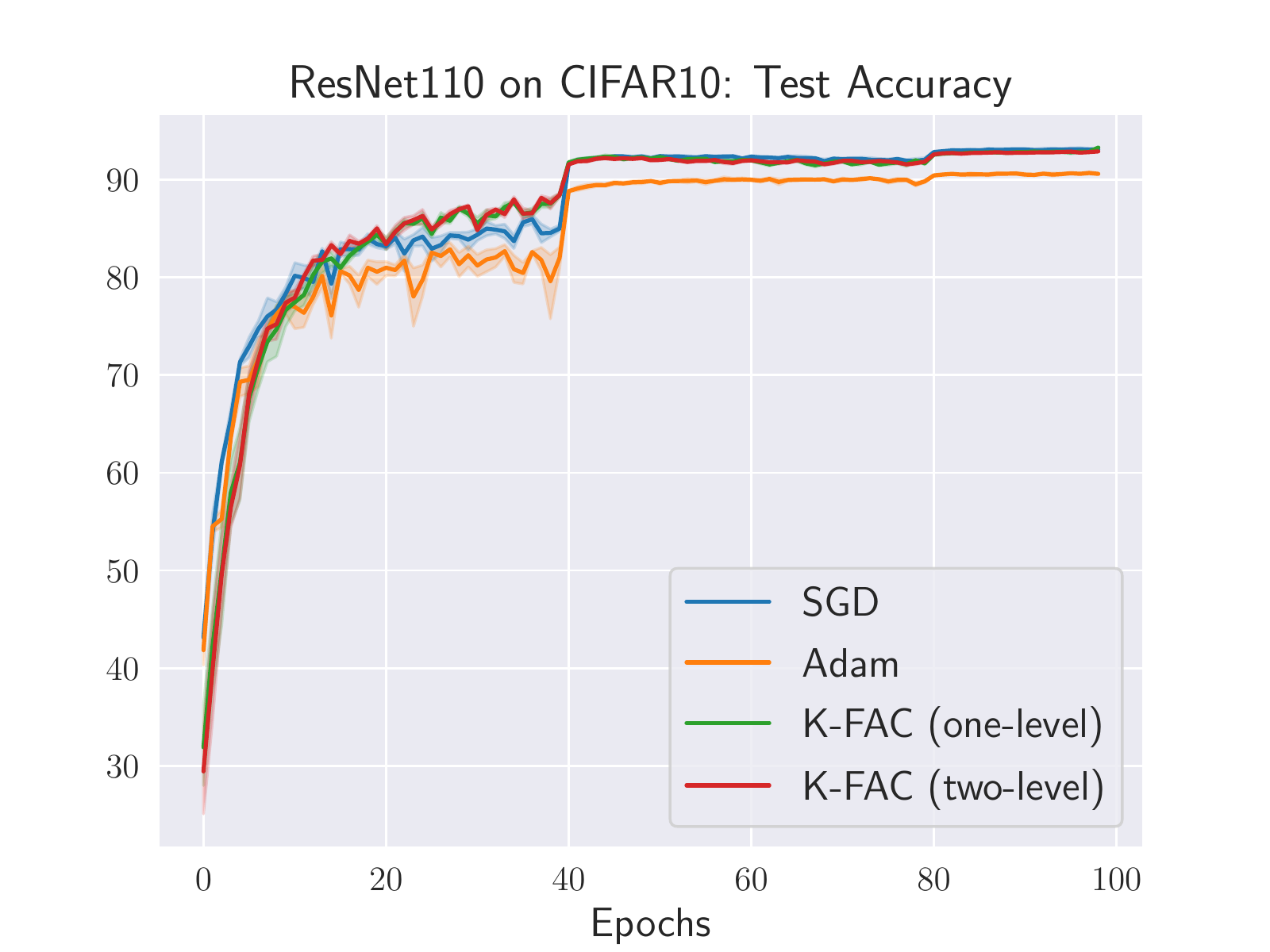}        
    \caption{Training and test accuracy per epoch for SGD, Adam, as well as one- and two-level K-FAC, when training ResNet110 on CIFAR10.}
\end{figure}

%% file: main.bbl
\begin{thebibliography}{34}
\providecommand{\natexlab}[1]{#1}
\providecommand{\url}[1]{\texttt{#1}}
\expandafter\ifx\csname urlstyle\endcsname\relax
  \providecommand{\doi}[1]{doi: #1}\else
  \providecommand{\doi}{doi: \begingroup \urlstyle{rm}\Url}\fi

\bibitem[Adolphs et~al.(2019)Adolphs, Kohler, and
  Lucchi]{adolphs2019ellipsoidal}
Leonard Adolphs, Jonas Kohler, and Aurelien Lucchi.
\newblock Ellipsoidal trust region methods and the marginal value of hessian
  information for neural network training.
\newblock \emph{arXiv preprint arXiv:1905.09201}, 2019.

\bibitem[Amari(1998)]{amari1998natural}
Shun-Ichi Amari.
\newblock Natural gradient works efficiently in learning.
\newblock \emph{Neural computation}, 10\penalty0 (2):\penalty0 251--276, 1998.

\bibitem[Arora et~al.(2018)Arora, Cohen, and Hazan]{arora2018optimization}
Sanjeev Arora, Nadav Cohen, and Elad Hazan.
\newblock On the optimization of deep networks: Implicit acceleration by
  overparameterization.
\newblock \emph{arXiv preprint arXiv:1802.06509}, 2018.

\bibitem[Ba et~al.(2016)Ba, Grosse, and Martens]{ba2016distributed}
Jimmy Ba, Roger Grosse, and James Martens.
\newblock Distributed second-order optimization using kronecker-factored
  approximations, 2016.

\bibitem[Bollapragada et~al.(2018)Bollapragada, Mudigere, Nocedal, Shi, and
  Tang]{bollapragada2018progressive}
Raghu Bollapragada, Dheevatsa Mudigere, Jorge Nocedal, Hao-Jun~Michael Shi, and
  Ping Tak~Peter Tang.
\newblock A progressive batching l-bfgs method for machine learning.
\newblock \emph{arXiv preprint arXiv:1802.05374}, 2018.

\bibitem[Botev et~al.(2017)Botev, Ritter, and Barber]{botev2017practical}
Aleksandar Botev, Hippolyt Ritter, and David Barber.
\newblock Practical gauss-newton optimisation for deep learning.
\newblock \emph{arXiv preprint arXiv:1706.03662}, 2017.

\bibitem[Cartis et~al.(2011)Cartis, Gould, and Toint]{cartis2011adaptive}
Coralia Cartis, Nicholas~IM Gould, and Philippe~L Toint.
\newblock Adaptive cubic regularisation methods for unconstrained optimization.
  part ii: worst-case function-and derivative-evaluation complexity.
\newblock \emph{Mathematical programming}, 130\penalty0 (2):\penalty0 295--319,
  2011.

\bibitem[Chapelle and Erhan(2011)]{chapelle2011improved}
Olivier Chapelle and Dumitru Erhan.
\newblock Improved preconditioner for hessian free optimization.
\newblock In \emph{NIPS Workshop on Deep Learning and Unsupervised Feature
  Learning}, volume 201. Sierra Nevada Spain, 2011.

\bibitem[Dolean et~al.(2012)Dolean, Nataf, Scheichl, and
  Spillane]{dolean2012analysis}
Victorita Dolean, Fr{\'e}d{\'e}ric Nataf, Robert Scheichl, and Nicole Spillane.
\newblock Analysis of a two-level schwarz method with coarse spaces based on
  local dirichlet-to-neumann maps.
\newblock \emph{Computational Methods in Applied Mathematics}, 12\penalty0
  (4):\penalty0 391--414, 2012.

\bibitem[Duchi et~al.(2011)Duchi, Hazan, and Singer]{duchi2011adaptive}
John Duchi, Elad Hazan, and Yoram Singer.
\newblock Adaptive subgradient methods for online learning and stochastic
  optimization.
\newblock \emph{Journal of machine learning research}, 12\penalty0 (7), 2011.

\bibitem[George et~al.(2018)George, Laurent, Bouthillier, Ballas, and
  Vincent]{george2018fast}
Thomas George, C{\'e}sar Laurent, Xavier Bouthillier, Nicolas Ballas, and
  Pascal Vincent.
\newblock Fast approximate natural gradient descent in a kronecker factored
  eigenbasis.
\newblock In \emph{Advances in Neural Information Processing Systems}, pages
  9550--9560, 2018.

\bibitem[Grosse and Martens(2016)]{grosse2016kronecker}
Roger Grosse and James Martens.
\newblock A kronecker-factored approximate fisher matrix for convolution
  layers.
\newblock In \emph{International Conference on Machine Learning}, pages
  573--582, 2016.

\bibitem[Jacot et~al.(2018)Jacot, Gabriel, and Hongler]{jacot2018neural}
Arthur Jacot, Franck Gabriel, and Cl{\'e}ment Hongler.
\newblock Neural tangent kernel: Convergence and generalization in neural
  networks.
\newblock In \emph{Advances in neural information processing systems}, pages
  8571--8580, 2018.

\bibitem[Jolivet et~al.(2014)Jolivet, Hecht, Nataf, and
  Prud'Homme]{jolivet2014scalable}
Pierre Jolivet, Fr{\'e}d{\'e}ric Hecht, Fr{\'e}d{\'e}ric Nataf, and Christophe
  Prud'Homme.
\newblock Scalable domain decomposition preconditioners for heterogeneous
  elliptic problems.
\newblock \emph{Scientific Programming}, 22\penalty0 (2):\penalty0 157--171,
  2014.

\bibitem[Kingma and Ba(2014)]{kingma2014adam}
Diederik~P Kingma and Jimmy Ba.
\newblock Adam: A method for stochastic optimization.
\newblock \emph{arXiv preprint arXiv:1412.6980}, 2014.

\bibitem[Kohler and Lucchi(2017)]{kohler2017sub}
Jonas~Moritz Kohler and Aurelien Lucchi.
\newblock Sub-sampled cubic regularization for non-convex optimization.
\newblock \emph{arXiv preprint arXiv:1705.05933}, 2017.

\bibitem[Kunstner et~al.(2019)Kunstner, Hennig, and
  Balles]{kunstner2019limitations}
Frederik Kunstner, Philipp Hennig, and Lukas Balles.
\newblock Limitations of the empirical fisher approximation for natural
  gradient descent.
\newblock In \emph{Advances in Neural Information Processing Systems}, pages
  4156--4167, 2019.

\bibitem[LeCun et~al.(2012)LeCun, Bottou, Orr, and
  M{\"u}ller]{lecun2012efficient}
Yann~A LeCun, L{\'e}on Bottou, Genevieve~B Orr, and Klaus-Robert M{\"u}ller.
\newblock Efficient backprop.
\newblock In \emph{Neural networks: Tricks of the trade}, pages 9--48.
  Springer, 2012.

\bibitem[Martens(2010)]{martens2010deep}
James Martens.
\newblock Deep learning via hessian-free optimization.
\newblock In \emph{ICML}, volume~27, pages 735--742, 2010.

\bibitem[Martens(2014)]{martens2014new}
James Martens.
\newblock New insights and perspectives on the natural gradient method.
\newblock \emph{arXiv preprint arXiv:1412.1193}, 2014.

\bibitem[Martens and Grosse(2015)]{martens2015optimizing}
James Martens and Roger Grosse.
\newblock Optimizing neural networks with kronecker-factored approximate
  curvature.
\newblock In \emph{International conference on machine learning}, pages
  2408--2417, 2015.

\bibitem[Martens et~al.(2018)Martens, Ba, and Johnson]{martens2018kronecker}
James Martens, Jimmy Ba, and Matt Johnson.
\newblock Kronecker-factored curvature approximations for recurrent neural
  networks.
\newblock In \emph{International Conference on Learning Representations}, 2018.

\bibitem[Napov and Notay(2012)]{napov2012algebraic}
Artem Napov and Yvan Notay.
\newblock An algebraic multigrid method with guaranteed convergence rate.
\newblock \emph{SIAM journal on scientific computing}, 34\penalty0
  (2):\penalty0 A1079--A1109, 2012.

\bibitem[Nocedal and Wright(2006)]{nocedal2006numerical}
Jorge Nocedal and Stephen Wright.
\newblock \emph{Numerical optimization}.
\newblock Springer Science \& Business Media, 2006.

\bibitem[Pascanu and Bengio(2013)]{pascanu2013revisiting}
Razvan Pascanu and Yoshua Bengio.
\newblock Revisiting natural gradient for deep networks.
\newblock \emph{arXiv preprint arXiv:1301.3584}, 2013.

\bibitem[Paszke et~al.(2019)Paszke, Gross, Massa, Lerer, Bradbury, Chanan,
  Killeen, Lin, Gimelshein, Antiga, et~al.]{paszke2019pytorch}
Adam Paszke, Sam Gross, Francisco Massa, Adam Lerer, James Bradbury, Gregory
  Chanan, Trevor Killeen, Zeming Lin, Natalia Gimelshein, Luca Antiga, et~al.
\newblock Pytorch: An imperative style, high-performance deep learning library.
\newblock In \emph{Advances in neural information processing systems}, pages
  8026--8037, 2019.

\bibitem[Pearlmutter(1994)]{pearlmutter1994fast}
Barak~A Pearlmutter.
\newblock Fast exact multiplication by the hessian.
\newblock \emph{Neural computation}, 6\penalty0 (1):\penalty0 147--160, 1994.

\bibitem[Pilanci and Wainwright(2017)]{pilanci2017newton}
Mert Pilanci and Martin~J Wainwright.
\newblock Newton sketch: A near linear-time optimization algorithm with
  linear-quadratic convergence.
\newblock \emph{SIAM Journal on Optimization}, 27\penalty0 (1):\penalty0
  205--245, 2017.

\bibitem[Schraudolph(2002)]{schraudolph2002fast}
Nicol~N Schraudolph.
\newblock Fast curvature matrix-vector products for second-order gradient
  descent.
\newblock \emph{Neural computation}, 14\penalty0 (7):\penalty0 1723--1738,
  2002.

\bibitem[Tang et~al.(2009)Tang, Nabben, Vuik, and Erlangga]{tang2009comparison}
Jok~Man Tang, Reinhard Nabben, Cornelis Vuik, and Yogi~A Erlangga.
\newblock Comparison of two-level preconditioners derived from deflation,
  domain decomposition and multigrid methods.
\newblock \emph{Journal of scientific computing}, 39\penalty0 (3):\penalty0
  340--370, 2009.

\bibitem[Wang et~al.(2019)Wang, Zhou, Liang, and Lan]{wang2019stochastic}
Zhe Wang, Yi~Zhou, Yingbin Liang, and Guanghui Lan.
\newblock Stochastic variance-reduced cubic regularization for nonconvex
  optimization.
\newblock In \emph{The 22nd International Conference on Artificial Intelligence
  and Statistics}, pages 2731--2740. PMLR, 2019.

\bibitem[Xu et~al.(2019)Xu, Roosta, and Mahoney]{xu2019newton}
Peng Xu, Fred Roosta, and Michael~W Mahoney.
\newblock Newton-type methods for non-convex optimization under inexact hessian
  information.
\newblock \emph{Mathematical Programming}, pages 1--36, 2019.

\bibitem[Xu et~al.(2020)Xu, Roosta, and Mahoney]{xu2020second}
Peng Xu, Fred Roosta, and Michael~W Mahoney.
\newblock Second-order optimization for non-convex machine learning: An
  empirical study.
\newblock In \emph{Proceedings of the 2020 SIAM International Conference on
  Data Mining}, pages 199--207. SIAM, 2020.

\bibitem[Zhang et~al.(2019)Zhang, Li, Nado, Martens, Sachdeva, Dahl, Shallue,
  and Grosse]{zhang2019algorithmic}
Guodong Zhang, Lala Li, Zachary Nado, James Martens, Sushant Sachdeva, George
  Dahl, Chris Shallue, and Roger~B Grosse.
\newblock Which algorithmic choices matter at which batch sizes? insights from
  a noisy quadratic model.
\newblock In \emph{Advances in Neural Information Processing Systems}, pages
  8196--8207, 2019.

\end{thebibliography}
